\documentclass{article}

\usepackage{arxiv}

\usepackage[utf8]{inputenc}
\usepackage[numbers,comma,square]{natbib}
\usepackage[figuresleft]{rotating}
\usepackage{txfonts}%
\usepackage{multirow}
\usepackage{array}
\usepackage{xcolor}
\usepackage[hidelinks]{hyperref}
\usepackage{algorithm}
\usepackage[noend]{algpseudocode}

\newcolumntype{L}[1]{>{\raggedright\let\newline\\\arraybackslash\hspace{0pt}}m{#1}}
\newcolumntype{C}[1]{>{\centering\let\newline\\\arraybackslash\hspace{0pt}}m{#1}}
\newcolumntype{R}[1]{>{\raggedleft\let\newline\\\arraybackslash\hspace{0pt}}m{#1}}

\definecolor{myML}{HTML}{08519C}
\definecolor{myOR}{HTML}{3182BD}
\definecolor{myNET}{HTML}{6BAED6}
\definecolor{myPS}{HTML}{BDD7E7}
\definecolor{mySTAT}{HTML}{EFF3FF}

\title{Analyzing Flight Delay Prediction Under Concept Drift}

\author{
	Lucas~Giusti \\
	CEFET/RJ\\
	\And
	Leonardo~Carvalho \\
	CEFET/RJ\\
	\And
	Antonio~Tadeu~Gomes \\
	LNCC\\
	\texttt{atagomes@lncc.br} \\
	\AND
	Rafaelli~Coutinho \\
	CEFET/RJ\\
	\texttt{rafaelli.coutinho@cefet-rj.br} \\
	\And
	Jorge~Soares \\
	CEFET/RJ\\
	\texttt{jorge.soares@cefet-rj.br} \\
	\And
	Eduardo~Ogasawara\\
	CEFET/RJ\\
	\texttt{eogasawara@ieee.org} \\
}

\begin{document}
\maketitle

\begin{abstract}
Flight delays impose challenges that impact any flight transportation system. Predicting when they are going to occur is an important way to mitigate this issue. However, the behavior of the flight delay system varies through time. This phenomenon is known in predictive analytics as concept drift. This paper investigates the prediction performance of different drift handling strategies in aviation under different scales (models trained from flights related to a single airport or the entire flight system). Specifically, two research questions were proposed and answered: (i) How do drift handling strategies influence the prediction performance of delays? (ii) Do different scales change the results of drift handling strategies? In our analysis, drift handling strategies are relevant, and their impacts vary according to scale and machine learning models used.

\end{abstract}

\section{Introduction}

Delay is one of the most critical indicators for flight transportation systems. Flight delays impose a challenge that impacts any flight transportation system. In the United States (US), it is estimated that a 10\% decrease in flight delays would mean an US\$ 8 billion (year base 2013) increase in Gross Domestic Product (GDP) \citep{peterson_economic_2013}. In this context, the prediction of delayed flights may be an essential tool for effectively addressing this problem.

The development of machine learning models makes it possible to identify potentially delayed flights or critical periods before happening, enabling better planning. For that reason, many predictive models have been developed to achieve the task \citep{rong_prediction_2015,kim_deep_2016,yu_flight_2019}. Commonly, flight data is combined with weather information from departure and arrival locations to help predict flight delays \citep{du_delay_2018,wu_modelling_2019}.

From the machine learning point of view, predicting delay may be a regression or classification task. In the former, the goal is to predict the amount of time (usually measured in minutes) a flight will delay. In the latter, the goal is to predict whether the flight is going to delay \citep{kim_deep_2016,gui_flight_2020}. The literature specialized in flight delay prediction provides many different models that have been developed with good results for both tasks. Specifically, for classification, which is the focus of the present study, random forest and deep recurrent neural network models have shown promising results in the US and China datasets \citep{rong_prediction_2015,yu_flight_2019}.

Large flight systems (such as the US, China, Europe, and Brazil) have challenges that impact flight delays. The relation of delays with input variables, such as destination or weather, may vary according to time and space \citep{sternberg_analysis_2016}. Thus, the proportion of delays may vary from time to time. Such a variation may occur due to punctual events such as storms and strikes. Other variations are disruptive, such as the FIFA World Cup in 2014. It led to an increase in the airports’ capacity throughout most of the Brazilian Flight System and a significant change in flights and passengers. 

Nevertheless, even when there is no perceived change in the size of the flight system, the relationship between system variables and delays may vary. These relationship variations lead to a scenario called \textit{concept drift} \citep{iwashita_overview_2019}. When the variables change but do not interfere with how delay occurs, there is no concept drift. Generally, a concept drift is a (statistically significant) difference between the joint probability of input and output variables observed in different dataset samples. 

Previous studies indicate that concept drift may impact predictive models \citep{gama_survey_2014,webb_characterizing_2016,iwashita_overview_2019}. Some studies have tested algorithms that retrain the aviation models if drift is detected or used algorithms that may adapt to concept drift, like recurrent neural networks \citep{khamassi_drift_2014,pesaranghader_fast_2016,kim_deep_2016}. Moreover, the amount of data used to train each model (so-called \textit{scale}) varies across past studies: models trained from flights related to a single airport or the entire flight system. To the best of our knowledge, no study considered different system scales.

This paper investigates the prediction performance of different drift handling strategies in aviation under different scales. This paper considers two different scales: \textit{system-based} ($SB$) and \textit{airport-based} ($AB$). In ($SB$), all airports in the flight system are considered together. Conversely, in $AB$, each airport is taken into account separately. Specifically, two research questions were proposed and answered: (i) How do drift handling strategies influence the prediction performance of delays? (ii) Do different scales change the results of drift handling strategies? These questions are answered while studying the Brazilian flight system dataset. It is an integrated database containing flight operations data provided by the Brazilian National Civil Aviation Agency (ANAC) \citep{anac_brazilian_2017} and airport weather data provided by Automated Surface Observing Systems (ASOS) \citep{asos_automated_2019}. The main findings of the present work were the following:
\begin{itemize}
	\item When the frequency of drifts is sufficiently high (as is the case with the dataset used in this study), retraining machine learning models offer better models than training only once;
	\item Considering a single airport or all airports as the system's scale has little influence on the performance of drift detection;
	\item Both less- and highly frequent drifts negatively affect the performance of drift prediction;
	\item Training for drift detection with data from larger periods (months or years) leads to better drift predictions because it accounts for seasonality---to the best of our knowledge, related work to date has only considered concise periods (hours) for drift predictions;
	\item Choosing a classifier highly depends on the system's scale;
	\item Accuracy is not the only important prediction performance metric to be considered in concept drift studies, which is an important shortcoming of related work.
\end{itemize}

Besides this introduction, this paper is organized as follows. Section \ref{sec:background} presents the general background for delay prediction and concept drift. Section \ref{sec:related_works} presents the related work. Section \ref{sec:methodology} discusses the methodology used for drift analysis over $SB$ and $AB$. Section \ref{sec:exp_evaluation} presents main results. Finally, Section \ref{sec:conclusion} presents some concluding remarks and points out future work.

\section{Background}
\label{sec:background}
The background is divided into two parts. Section \ref{subsec:delay} presents flight delays prediction using machine learning. Section \ref{subsec:drift} presents concept drift, including drift detection and handling. 

\subsection{Flight Delays Prediction Using Machine Learning}
\label{subsec:delay}

Flight delay is a measure of the actual departure or arrival time minus their respective expected time. For the classification task of predicting whether a delay will occur or not, a threshold is set up to establish this binary variable. Most studies use a threshold of 15 minutes. It indicates that any flight 15 minutes or more late is marked as delayed \citep{gui_flight_2020,guleria_multi-agent_2019,kim_deep_2016,sternberg_analysis_2016}. 

Many studies have been conducted and reported good results in the classification task \citep{belcastro_using_2016,moreira_evaluating_2018}. In this context, Random Forests ($RF$) and Neural Networks ($NN$) achieved better $accuracy$. An $NN$ is a bio-inspired computational approach that performs the processing of information through neurons that are connected through synapses \citep{han_data_2011}. Specifically, Multi-Layer Perceptrons associated with Long Short Term Memory Recurrent Neural Network (LSTM-RNN) and Deep Multilayer Perceptrons seem to show good results during prediction \citep{gui_flight_2020,kim_deep_2016,rebollo_characterization_2014}. Due to its interpretability, Naive Bayes ($NB$) is commonly included in the studies, as it encompasses a baseline method. $NB$ is a statistical classifier that can predict the probability of a tuple belonging to a particular class.

\begin{figure}[!ht]
	\centering
	\includegraphics[width=.7\textwidth]{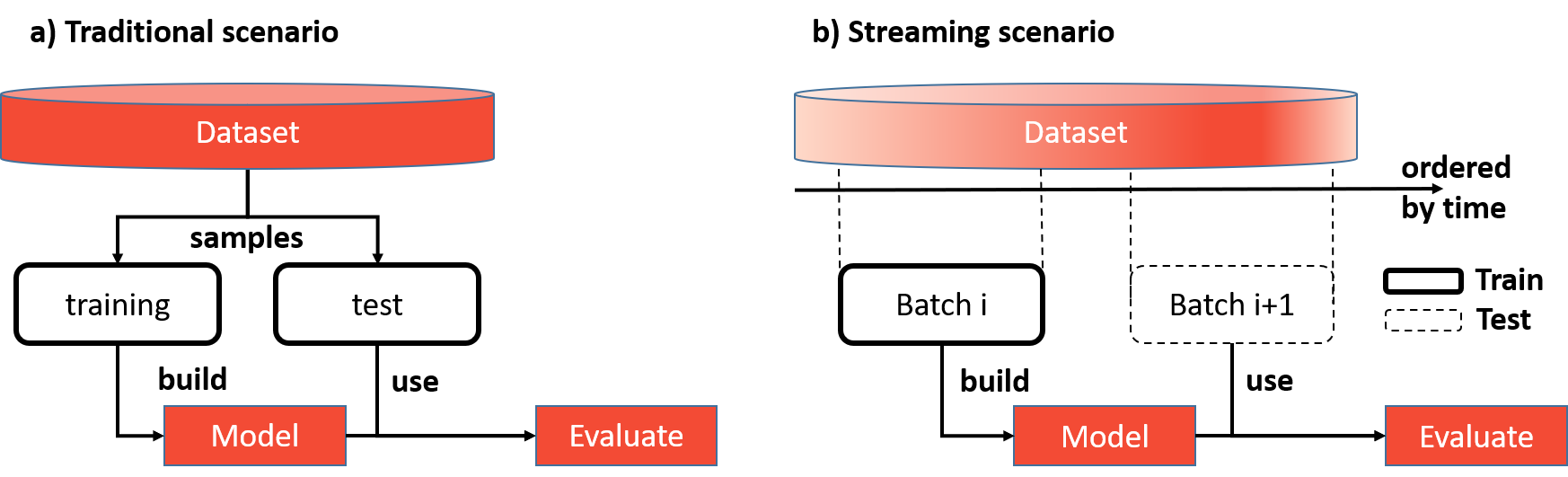}
	\caption{Characterization of training and testing during classification: (a) traditional scenario; (b): streaming scenario}
	\label{fig:training}
\end{figure}

In a traditional classification problem, a dataset is separated into training and test sets. The model is built using the training set. For that, it is common to partition the training set using cross-validation to do hyperparameters optimization. Once the model is built, it is later evaluated using a test set \citep{han_data_2011}. This traditional approach is depicted in Figure \ref{fig:training}.a. 

However, for flight delay prediction, the time dimension is relevant. Flights occur continuously each day as a streaming data source. Even when the entire streaming is stored in a single dataset, the time stamp of the flight events needs to be considered. It means that training should occur using past data to predict more recent data. It is depicted in Figure \ref{fig:training}.b, where the $i$-th batch (training data) is used to build a model for further evaluation with more recent data at the next batch ($i+1$) (test data). It is worth mentioning that the $i$-th batch corresponds to a sample of the dataset in the time interval associated with $i$. While studying concept drift, such methodology is mandatory \citep{iwashita_overview_2019}. 

Finally, metrics are used to evaluate the prediction performance of built models. Formally, given two classes (positive and negative): positive tuples corresponding to delays and negative tuples for the ones without it. $P$ is the number of positive tuples, and $N$ is the number of negative tuples. The class of the test set is compared to the class predicted by the built model, getting: True Positives ($TP$), True Negatives ($TN$), False Positives ($FP$), and False Negatives ($FN$) \citep{han_data_2011,moreira_evaluating_2018}. From these measures, it is possible to compute the most widely used metrics: $accuracy$ ($\frac{TP+TN}{P+N}$), $precision$ ($\frac{TP}{TP+FP}$), $recall$ ($\frac{TP}{TP+FN}$), and $f_1$ ($\frac{2 \times precision \times recall}{precision + recall}$).

\subsection{Concept Drift}
\label{subsec:drift}

Consider a classification problem, such that a set of input variables $X$ is used to predict a class label $Y$. One of the main challenges when creating machine learning models is handling concept drift. It refers to a significant change in data distribution that interferes with the relation between the output class $Y$ and input variables $X$. Formally, a concept at a time $i$ is defined as the probability of the joint distribution $\chi$ of $X$ and $Y$. It is described in Equation \ref{eq:concept}. A concept drift between time $i$ and $j$ is defined as a difference (with statistical significance) between the probabilities $p_i(\chi)$ and $p_j(\chi)$. It is described in Equation \ref{eq:drift} \citep{webb_characterizing_2016}.

\begin{equation}\label{eq:concept} 
	concept_i = p_i(\chi) = p_i(X, Y)
\end{equation}

\begin{equation}\label{eq:drift} 
	p_i(\chi) \neq p_j(\chi) 
\end{equation}

\begin{figure}[!ht]
	\centering
	\includegraphics[width=.7\textwidth]{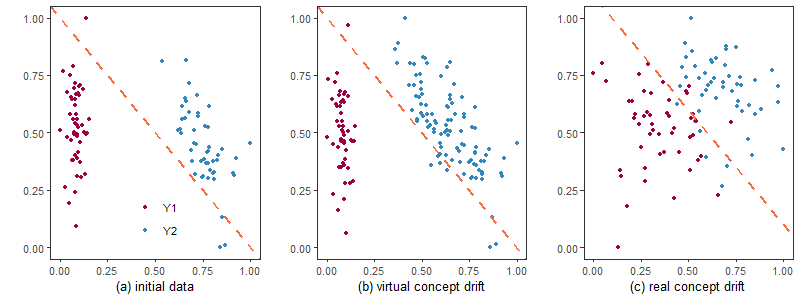}
	\caption{Examples of types of drifts: (a) initial dataset; (b) virtual concept drift from (a); (c) real concept drift from (a)}
	\label{fig:drifts}
\end{figure}

The drifts can also be classified as real or virtual drifts \citep{iwashita_overview_2019}. Specifically, real concept drifts are defined by changes in the posterior probabilities $p(Y|X)$, which is commonly related to the class boundaries. Conversely, virtual concept drifts happen whenever the conditional probability $p(X|Y)$ changes but $p(Y|X)$ holds \citep{iwashita_overview_2019, lu_learning_2018, hoens_learning_2012}. A visual example is shown in Figure \ref{fig:drifts}.

Regarding flight delay, there can be relevant changes in the proportion of delayed flights. Some periods are more critical than others \citep{gui_flight_2020,sternberg_analysis_2016}. For that reason, detecting and handling drifts is a relevant subject. Drift detection refers to the task of identifying concept drifts. It enables a specific action to avoid increasing errors in online learning systems after drift is observed \citep{iwashita_overview_2019,lu_learning_2018,webb_characterizing_2016}. There are two main categories of drift detection \citep{lu_learning_2018}: (i) data distribution and (ii) error rate. Data distribution-based methods use statistical inference and analysis of feature distribution to detect significant output class proportion changes concerning its input variables. Error rate methods use machine learning algorithms and indicate a drift based on the error rate of prediction results. The detection of drifts can be based solely on data distribution, the error rate of predictions, or both. 

Consider a dataset or a streaming dataset ($D$) partitioned into batches (time intervals of the same size). $D_1$ and $D_n$ correspond to the first and last batches of $D$, respectively. Yet, a batch sequence $b$ at time $i$ is formally defined as $seq_{i,b}(D) = \textless D_{i-b+1}, \cdots, D_{i}\textgreater$. Indeed, a Batch Sequence Size ($BSS$) equals $b$ establishes a sliding window to explore all batch sequences of size $b$ present in $D$. It can be used to target both the detection and handling of concept drifts. It can be formalized as $sw_b(D)$. It corresponds to a matrix $W$ of size ($n - b + 1$) by $b$. Each line $w_i$ in $W$ is the $i$-th $BSS$ $b$ in $D$. Given $W = sw_b(D)$, $\forall w_i \in W$, $w_i = seq_{i,b}(D)$ \citep{iwashita_overview_2019,lu_learning_2018}. 

\begin{figure}[!ht]
	\centering
	\includegraphics[width=.9\textwidth]{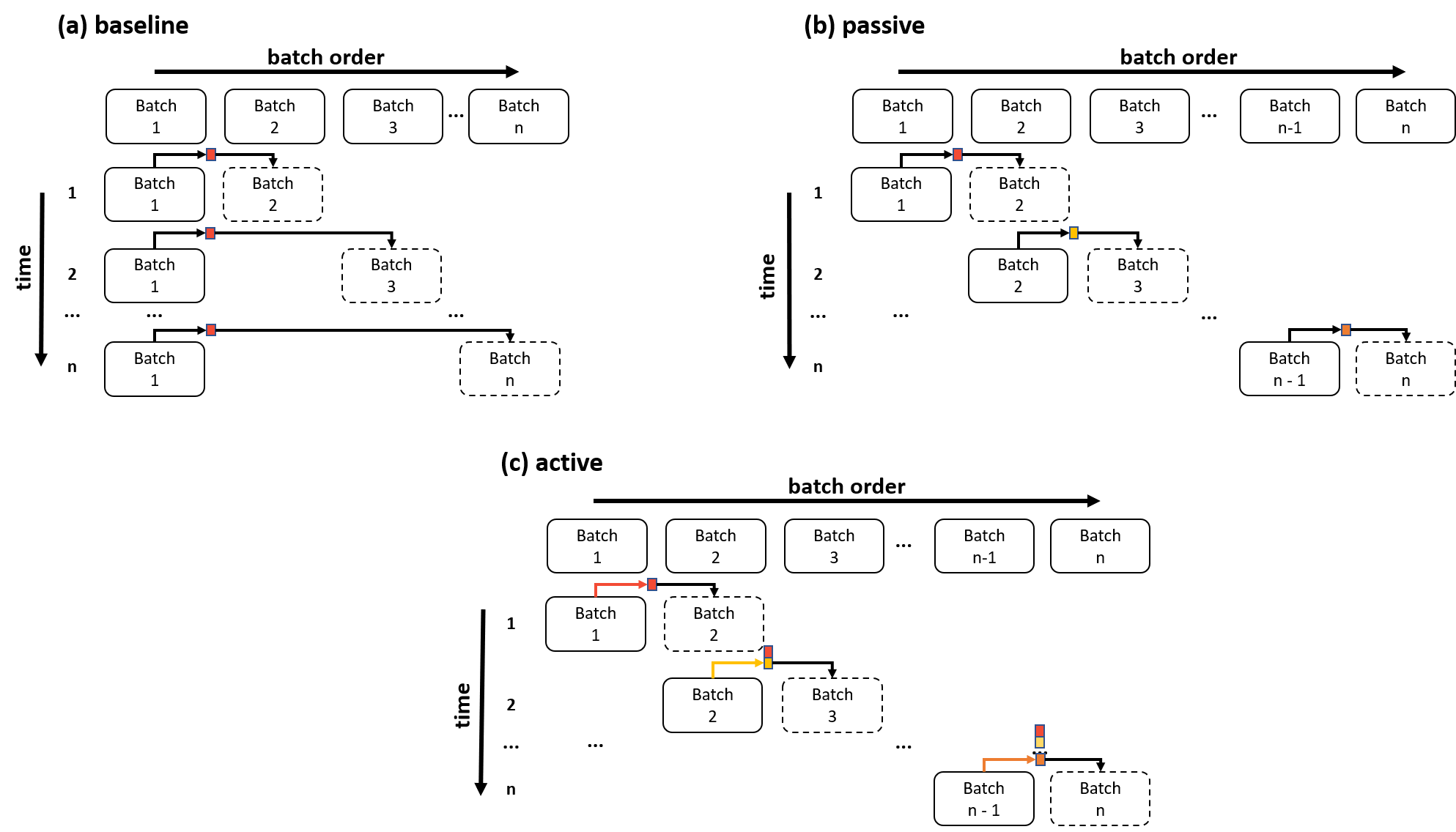}
	\caption{Drift handling strategies considering a $BSS$ equal to one ($b = 1$). (a): $baseline$; (b): $passive$; (c): $active$}
	\label{fig:drift_detection_handling}
\end{figure}

From these concepts, it is possible to define three strategies to address concept drift: (i) $baseline$; (ii) $passive$; (iii) $active$. In the $baseline$ strategy, a model is built using the first batch. The trained model is continuously used. When drift occurs, no action is done, and the trained model might increase its error during the prediction of newer batches. Considering $BSS$ equals one ($b = 1)$, it corresponds to Figure \ref{fig:drift_detection_handling}.a, where the first batch ($1$) is used for training a model (indicated as a red square) for predicting all other batches ($2$ to $n$). 

In a $passive$ strategy, it is assumed that drift occurs all the time. Thus, considering again a $BSS$ equals one $b = 1$, batch $i$ is used for training the model to predict batch $i+1$. This scenario corresponds to Figure \ref{fig:drift_detection_handling}.b. Models are constantly updated (they are presented in different colors). The drawback of this approach is that it might retrain models, even if no drift occurred in the dataset \citep{iwashita_overview_2019, gama_survey_2014}.

Finally, in the $active$ strategy, whenever a new batch is introduced, drift detection is applied. If no drift is detected, the previously trained model is still used. However, if drift is detected, a new model is built using previous batches. Figure \ref{fig:drift_detection_handling}.c depicts this scenario for $b = 1$. If a drift occurred between batches two and three, a new model (presented in orange) is used. Otherwise, the previous model (presented in red) is preserved. In this strategy, two extreme scenarios may occur. The same model can be used from the first batch to the last one, resembling the baseline strategy. The difference is that the decision is because no drift was observed. Conversely, continuous retraining may occur between each pair of batches, resembling the $passive$ strategy. Again, such a decision is based on whether drift is observed whenever a new batch is introduced \citep{iwashita_overview_2019, gama_survey_2014}. 

\section{Related Work}
\label{sec:related_works}

Past literature on delay prediction is extensive \citep{carvalho_relevance_2020}. We have carried out a systematic mapping study to identify relevant data science studies regarding flight delay prediction\footnote{Search string used: ( ``flight delay'' ) {\sc and} ( ``classification'' {\sc or} ``regression'' {\sc or} ``prediction'')}. There were deemed only papers (Journals and Conferences) entirely written in English. The query was submitted to the Scopus database (considering titles, keywords, and abstracts) in December 2020 and returned 141 papers. Additionally, two studies were added through snowballing.

Regarding training and testing, the majority of studies uses traditional scenario (Figure \ref{fig:training}.a) \citep{chen_fuzzy_2008,khanmohammadi_systems_2014,alonso_predicting_2015}. However, this approach does not consider the possible drifts that are usually present on flight data. Studies that considered streaming approaches were deeply investigated, particularly those that contained drift handling strategies \citep{kim_deep_2016, khamassi_drift_2014, pesaranghader_fast_2016, munoz_hernandez_real-time_2019,wang_airport_2019,ai_deep_2019,gui_flight_2020}. In fact, only the work of \citet{kim_deep_2016}, \citet{khamassi_drift_2014}, \citet{pesaranghader_fast_2016} investigated the binary classifier problem. 

\citet{kim_deep_2016} used Deep Neural Networks predict flight delays in the US flights using a $BSS$ of seven and nine days before prediction. An additional Deep Recurrent Neural Network was trained to classify the critical delay status of each day. Since the scale used was of flight routes, these predictions were added to individual flight routes data and used as input in a Deep Neural Network. Although the $accuracy$ was above 80\% in most experiments, key metrics such as $f_1$, $precision$, and $recall$ were not reported. As no specific strategy was used to identify or deal with drifts actively, we considered this approach as a passive strategy.

\citet{khamassi_drift_2014} proposed a new error active distance-based approach for drift detection and monitoring, named EDIST. Specifically, EDIST compares new data with existing data and retrains the model if significant changes are found in the distribution. The classifier used in this case was the Hoeffding Trees, and $BSS$ was statistically adaptive. Active strategies of the Drift Detection Method (DDM) and the Early DDM (EDDM) are both error-based methods. They were implemented for comparison reasons. A baseline strategy of training with the first batch and predicting the remaining data was tested as well. They have applied these techniques to many synthetic and real-world datasets (including the US flight dataset). However, only the accuracy was used to report prediction performance. 

Finally, \citet{pesaranghader_fast_2016} tested many drift detection techniques with Naive Bayes and Hoeffding Trees. Specifically, their work proposes using the Hoeffding Inequality Theorem to test the difference in the probability of a given class between two $BSS$ of 25 cases, characterizing an active strategy. DDM, EDDM, Adaptive Sliding Window (ADWIN), Hoeffding Drift Detection Method (HDDM), and Fast HDDM (FHDDM) active strategies were implemented for comparison reasons. ADWIN implements statistically adaptive batch sizes, defined whenever a drift is detected by average comparison. HDDM uses the Hoeffding inequality to compare distributions of batches. Finally, the proposed FHDDM uses Hoeffding inequality to compare errors from batches and thus detect drifts. The reported $accuracy$ was around 65\% for all experiments with aviation data. Fast Hoeffding Trees with Adaptive Windowing showed the best results for the US flight datasets.

To the best of our knowledge and considering the systematic mapping presented here, no study compared active and passive drift handling strategies for delay prediction in aviation. Moreover, the influence of the scale of the data used to train the models was never investigated as an important factor for delay prediction. Besides accuracy, no other prediction performance indicators were reported as well.

\section{Methodology}
\label{sec:methodology}

This paper aims to study drift handling strategies. Specifically, three different $active$ strategies and one $passive$ strategy were investigated. A second goal is to investigate drift handling strategies under the influence of different scales of training data: System-Based ($SB$) and Airport-Based ($AB$). In other words, how the scales interfere with drift handling strategies.

Pseudo-code \ref{alg:methododology} describes our methodology for flight delay prediction with concept drift. It requires seven parameters: $D$, $airport$, $mlm$, $t$, $b$, $dd$, and $dh$. The parameter $D$ corresponds to the input dataset. The parameter $airport$ identifies a single airport if the $AB$ scale is used, or $nil$ if the $SB$ is used. Parameter $mlm$ corresponds to the machine learning method. Parameter $t$ is related to the time, and $b$ corresponds to the size of the batch sequence. Finally, $dd$ and $dh$ correspond to the drift detection method and drift handling strategy, respectively. These parameters are described in Table \ref{tbl:parameters}.

\floatname{algorithm}{Pseudo-code}
\begin{algorithm}[!ht]
	\begin{algorithmic}[1]
		\Function{$methodology$}{$D, airport, mlm, t, b, dd, dh$}
		\State $D \gets preprocess(D, airport)$
		\State $D_i \gets selectBatchesTrain(D, t, b)$
		\State $D_j \gets selectBatchesTrain(D, t-1, b)$
		\State $train \gets actDrift(dd, dh, D_i, D_j)$
		\If{$train$}
		\State $mdl \gets trainModel(mlm, D_i)$
		\State $storeModel(mdl, airport, mlm, dd, dh)$
		\Else
		\State $mdl \gets loadModel(airport, mlm, dd, dh)$
		\EndIf
		\State $D_{t+1} \gets selectBatchTest(D, t+1)$
		\State $results \gets test(mdl, D_{t+1})$
		\State return $results$
		\EndFunction
	\end{algorithmic}
	\caption{General methodology}
	\label{alg:methododology}
\end{algorithm}

\begin{table}[!ht]
	\centering
	\caption{Parameters used}
	\begin{tabular}{C{1.5cm} C{13cm}}
		\hline
		\textbf{Name} & \textbf{Description [values]}\\
		\hline
		$D$ & Integrated dataset \citep{teixeira_brazilian_2020} containing all Brazilian flights (ANAC) with weather records (ASOS) [from 2000 to 2018] \\
		$airport$ & Airport code for the ten most significant airports in Brazil for $AB$ analysis; or $nil$, for $SB$ analysis [$nil$, $SBBR$, $SBPA$, $SBSV$, $SBGL$, $SBCT$, $SBKP$, $SBGR$, $SBCF$, $SBRJ$, $SBSP$] \\
		$mlm$ & Machine learning models: Naive Bayes ($NB$), Neural Networks ($NN$), Random Forest ($RF$) [$NB$, $NN$, $RF$] \\
		$t$ & Time (yearly based time slices) [from 2003 to 2017] \\
		$b$ & $BSS$ for training [from 1 to 3]\\
		$dd$ & Drift detection [$mean$, $variance$, $mean/variance$]\\
		$dh$ & Drift handling [$baseline$, $active$, $passive$]\\
		\hline
	\end{tabular}
	\label{tbl:parameters}
\end{table}

The first step of Pseudo-code \ref{alg:methododology} does the data preprocessing (function $preprocess$). If $airport$ is different from $nil$, the dataset $D$ is filtered for that airport. Otherwise, it studies the entire $SB$. All steps of data preprocessing are described in Section \ref{subsec:met_preprocessing}. Lines 3 and 4 prepare training batches $D_i$ and $D_j$ from the $BSS$ of $b$. $D_j$ is lagged one position to enable detection of concept drift. 

Line 5 computes the drift action. It considers the drift detection method ($dd$), drift handling strategy ($dh$), and the two batch sequences ($D_i$ and $D_j$). The mechanics of drift action are described in Section \ref{subsec:met_drift_action}. The output of drift action is assigned to $train$, indicating if training is required at time $t$. If required, a machine learning method is trained using the batch sequence $D_i$ (line 7). The training process is described in Section \ref{subsec:met_model_training}. The output is a trained model ($mdl$). The model is stored for the selected $airport$, $mlm$, $dd$, and $dh$. If no training is required, previously trained model $mdl$ is retrieved for the selected $airport$, $mlm$, $dd$, and $dh$.

Line 11 selects the batch for testing ($D_{t+1}$). The model $mdl$ is used to predict delays. The methodology returns the results of the prediction. From Pseudo-code \ref{alg:methododology}, the experimental setup is described in Section \ref{subsec:met_experimental_setup}.

\subsection{Preprocessing}
\label{subsec:met_preprocessing}

\begin{figure}[!ht]
	\centering
	\includegraphics[width=0.5\textwidth]{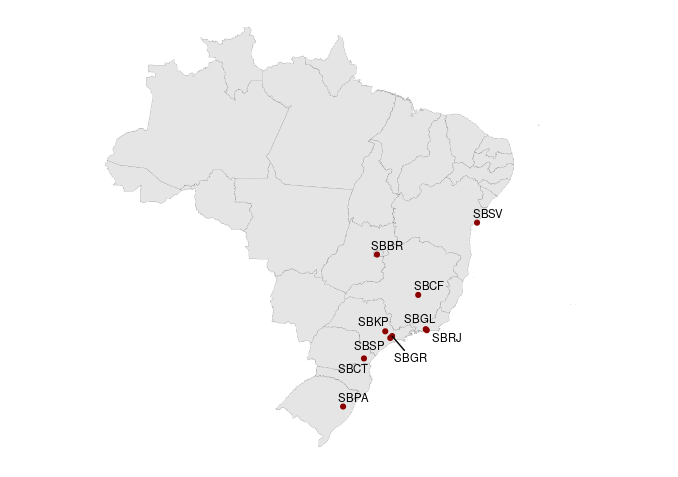}
	\caption{Location of ten main Brazilian airports}
	\label{fig:aiportsMap}
\end{figure}

The dataset used is the Brazilian Flights Dataset \citep{teixeira_brazilian_2020}. It is an integrated dataset containing ANAC’s flight operations \citep{anac_brazilian_2017} with ASOS's airport weather data \citep{asos_automated_2019}. It contains data from 2000 to 2018. For the sake of scale comparison, the data used in this paper is filtered for the top ten airports with the highest number of departing flights, and only domestic flights were evaluated. Graphical representation of the location of airports is shown in Figure \ref{fig:aiportsMap}, and descriptive information is shown in Table \ref{tab:airports}. For the $SB$ analysis, the filtered dataset is studied altogether. Conversely, for $AB$ analysis, each one of the top ten airports is studied separately.

In addition to filtering, three features were created as well. For destination, the state of each airport was used instead of the airport name. The week number within the year was created to perform statistical tests. According to international standards for flight delay, the binary variable for flight delay was set with a threshold of 15 minutes. The cases in which departure delay was higher than 24 hours or missing were considered errors and excluded from the analysis. The numeric attributes were normalized using the min-max technique.

\begin{table}[!ht]
	\centering
	\caption{Top ten Brazilian airports studied}
	\begin{tabular}{C{1.5cm} C{3cm} C{3.5cm}}
		\hline
		Code & City & State \\ \hline
		$SBBR$ & Brasília & Distrito Federal (DF) \\
		$SBSV$ & Salvador & Bahia (BA) \\
		$SBCT$ & São José dos Pinhais & Paraná (PR) \\
		$SBGL$ & Rio de Janeiro & Rio de Janeiro (RJ) \\
		$SBPA$ & Porto Alegre & Rio Grande do Sul (RS) \\
		$SBKP$ & Campinas & São Paulo (SP) \\
		$SBGR$ & Guarulhos & São Paulo (SP) \\
		$SBSP$ & São Paulo & São Paulo (SP) \\
		$SBCF$ & Belo Horizonte & Minas Gerais (MG) \\
		$SBRJ$ & Rio de Janeiro & Rio de Janeiro (RJ) \\ \hline
	\end{tabular}
	\label{tab:airports}
\end{table}

\subsection{Drift Action}
\label{subsec:met_drift_action}

All drift handling strategies were tested for the $SB$ and $AB$ scales. The years used as test sets varied from 2004 to 2018 for all methods to compare different $BSS$ for training ($b$ from $1$ to $3$). The minimum value for $BSS$ is one to include yearly seasonality in each trained classifier. It is important to create models that incorporate seasonal components of streaming data.

Specifically, we implemented the $baseline$, $active$, and $passive$ drift handling strategies. The $baseline$ corresponds to training using the first batch sequence to predict all other batches. The $passive$ strategy trained a model for each batch sequence to predict the next batch. Finally, the $active$ strategy compared the current training batch sequence with the previous one to detect if a drift occurred. In case of not having drift, the previously trained model is chosen. Otherwise, a new model is trained using the current training batch sequence. The prediction of the next batch is made using the chosen model. 

For $active$ strategies, we implemented three methods of drift detection based on data distribution analysis. They evaluate the occurrence of drifts according to the proportion of delays ($p(Y)$) presented in two consecutive batch sequences ($D_i$ and $D_{i-1}$). The three methods were (i) $mean$, (ii) $variance$, and (iii) $mean/variance$. In $mean$ and $variance$, respectively, the $mean$ and $variance$ of both training batches sequences were compared for a statistically significant difference. Finally, in $mean/variance$, a significant difference in either $mean$ or $variance$ indicates a drift. For these tests, the proportion of delays was aggregated in weeks and tested for normality with Kolmogorov Smirnov and Shapiro-Wilk tests \citep{yap_comparisons_2011}. For normal distributions, the mean t-test was used, and the F test was used for the $variance$. When distributions were not normal, the Wilcoxon test was used for the $mean$, and the Levene test was used for the $variance$. The p-value used was 0.05 for all tests.

\subsection{Model training} 
\label{subsec:met_model_training}

Three classification methods were used: $NB$, $RF$, and $NN$. Each technique was replicated five times for each predicted year and $BSS$, except for the deterministic Naive Bayes. A grid search was used to find the best hyperparameters for $NN$ and $RF$. 

The hyperparameters were the number of hidden neurons for $NN$, the number of randomly selected predictors for $RF$, and the score and smooth for $NB$. Ten-fold cross-validation was performed. The optimized hyperparameters were only computed for the first batch (2003). These hyperparameters were used in all trained models built by each classification method.

\subsection{Experimental Setup} 
\label{subsec:met_experimental_setup}

Considering the general methodology described in Pseudo-code \ref{alg:methododology} and parameters described in Table \ref{tbl:parameters}, Pseudo-code \ref{alg:experimental_evaluation} describes the entire concept drift analysis. It executes the methodology considering the cross-product for all possible values of the parameters. 

\begin{algorithm}[!ht]
	\begin{algorithmic}[1]
		\Function{$drift\_analysis$}{$D$}
		\State $A \gets \{nil, SBBR, SBPA, SBSV, SBGL, SBCT, SBKP, SBGR, SBCF, SBRJ, SBSP\}$
		\State $MLM \gets \{NB, NN, RF\}$
		\State $T \gets \{2003, \cdots, 2017\}$
		\State $B \gets \{1, 2, 3\}$
		\State $DD \gets \{mean, variance, mean/variance\}$
		\State $DH \gets \{baseline, passive, active\}$
		\State $results \gets nil$
		\For {\textbf{each}~ $a \in A, mlm \in MLM, t \in T, b \in B, dd \in DD, dh \in DH$}
		\State $results \gets results \cup methodology(D, a, mlm, t, b, dd, dh)$
		\EndFor
		\State return $results$
		\EndFunction
	\end{algorithmic}
	\caption{Experimental evaluation}
	\label{alg:experimental_evaluation}
\end{algorithm}

The data mining process was implemented in R for both preprocessing and machine learning methods \citep{han_data_2011,james_introduction_2013}. These are available as R packages (caret, nnet, randomForest, e1071, dplyr, PerformanceAnalysis). 

\section{Results}
\label{sec:exp_evaluation}

The experimental evaluation was driven to answer two research questions regarding flight delay prediction: (i) How do drift handling strategies influence the prediction performance (Section \ref{subsec:difference_dd})? (ii) Do different scales change the results of drift handling strategies (Section \ref{subsec:difference_bss})? 

After answering these research questions, Section \ref{subsec:discussion} discusses other important results. It compares our findings with other related works. The entire experimental evaluation was executed in one month on an i7 processor with $16$ cores with $128$GB of RAM and a Ubuntu $20.04$ operating system.

\subsection{How do drift handling strategies influence the prediction performance?}
\label{subsec:difference_dd}

The entire experimental evaluation execution considered the period from 2004 to 2018. In a yearly-based analysis, the total number of possible drifts is 15. The number of drifts is shown in Table \ref{tbl:drifts_full} for $SB$ analysis. They are presented according to $BSS$.

\begin{table}[!ht]
	\centering
	\caption{Number of detected drifts for $SB$ by $BSS$}
	\begin{tabular}{C{3cm} C{1.75cm} C{1.75cm} C{2.25cm}}
		\hline
		$BSS$ (training:test) & $mean$ & $variance$ & $mean/variance$ \\ \hline
		1:1 & 9 & 2 & 10  \\
		2:1 & 11 & 5 & 11 \\
		3:1 & 8 & 4 & 8  \\ \hline
	\end{tabular}
	\label{tbl:drifts_full}
\end{table}

Considering $mean$ and $mean/variance$ methods, the average number of drifts over the entire period was 9.33 and 9.66, respectively. These numbers correspond to 62.2\% and 64\% of all possible drifts, which indicates a high prevalence of drifts between batches for all $BSS$ sizes. Moreover, the $variance$ method showed an average of 3.66 (24\%), which indicates that this method is less sensitive to drifts in flight delays. 

\begin{figure}[!ht]
	\centering
	\includegraphics[width=0.9\textwidth]{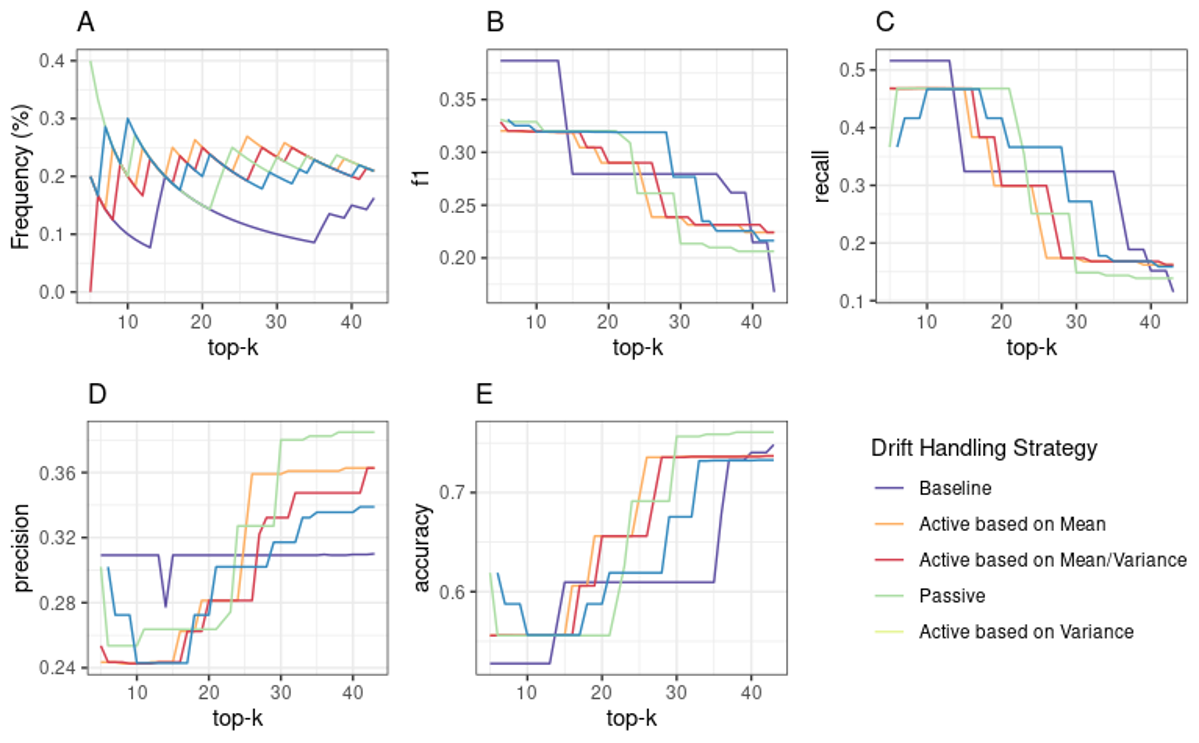}
	\caption{Results of top-$k$ analysis ranked by $f_1$. Frequency of each drift handling strategy (A). Median performance for $f_1$ (B), $recall$ (C), $precision$ (D), $accuracy$ (E).}
	\label{fig:topKStPerf}
\end{figure}

To identify the best drift handling strategies, we carried out an analysis of the top-$k$ best combinations ordered by $f_1$. Specifically, we computed the frequency of appearance of each strategy for each top-$k$ ($5 \leq k \leq 45$). The main results for frequency and performance are shown in Figure \ref{fig:topKStPerf}.a.

Regarding the frequency of each drift handling strategy, $active$ and $passive$ combinations were more frequent from the top 5. Specifically, $passive$ was the most frequent at the top 5 and tended to balance with other $active$ combinations as the $k$ increased. These results indicate that retrain machine learning models (whether $active$ or $passive$) can frequently offer better models than training only once in the beginning as in $baseline$.

Regarding the other metrics (Figures \ref{fig:topKStPerf}.b, \ref{fig:topKStPerf}.c, \ref{fig:topKStPerf}.d, and \ref{fig:topKStPerf}.e). For lower $k$ values (better ranked according $f_1$), $f_1$ and $recall$ values are higher, whereas $accuracy$ and $precision$ are lower. It indicates that models not targeting the majority class are better ranked. When $k$ gets higher, $f_1$ and $recall$ decrease, whereas $accuracy$ and $precision$ increase. It indicates that for higher $k$ (worse ranked according $f_1$), models that target the majority class are more common. 

\subsection{How may different scales of train data influence drift handling strategies?}
\label{subsec:difference_bss}

As shown in Table \ref{tbl:driftsAirports}, the average number of drifts for the $AB$ scale was very similar to $SB$. The most sensitive methods were also $mean$ and $mean/variance$, with 8.6 (57\% of all possible drifts) and 9.1 (61\% of all possible drifts). The $variance$ method was again less sensitive to drifts and showed a 4.4 average drifts (29.3\% prevalence). These results indicate that $AB$ scale may not influence drift detection when compared to $SB$ scale. 

\begin{table}[!ht]
	\centering
	\caption{Number of Drifts by each Airport, Drift Detection Method, and $BSS$.}
	\begin{tabular}{cccccccccc}
		\hline
		\multirow{2}{*}{Airport} & \multicolumn{3}{c}{$mean$} & \multicolumn{3}{c}{$variance$} & \multicolumn{3}{c}{$mean/variance$} \\ \cline{2-10}
		& 1 & 2 & 3 & 1 & 2 & 3 & 1 & 2 & 3 \\ \hline
		$SBBR$ & 8 & 8 & 9 & 3 & 4 & 5 & 8 & 8 & 9 \\
		$SBCF$ & 9 & 10 & 9 & 2 & 4 & 5 & 9 & 10 & 9 \\
		$SBCT$ & 8 & 8 & 8 & 1 & 4 & 5 & 8 & 8 & 8 \\
		$SBGL$ & 8 & 10 & 9 & 2 & 4 & 6 & 8 & 10 & 9 \\
		$SBGR$ & 12 & 8 & 6 & 6 & 7 & 5 & 12 & 10 & 6 \\
		$SBKP$ & 11 & 11 & 10 & 5 & 8 & 4 & 12 & 13 & 10 \\
		$SBPA$ & 10 & 10 & 9 & 5 & 5 & 7 & 12 & 11 & 11 \\
		$SBRJ$ & 6 & 4 & 3 & 3 & 5 & 3 & 8 & 6 & 3 \\
		$SBSP$ & 9 & 7 & 7 & 3 & 3 & 3 & 11 & 7 & 7 \\
		$SBSV$ & 10 & 11 & 10 & 4 & 5 & 6 & 10 & 11 & 11 \\
		Mean & $9.1 \pm 1.7$ & $8.7 \pm 2.2$ & $8.0 \pm 2.2$ & $3.4 \pm 1.6$ & $4.9 \pm 1.5$ & $4.9 \pm 1.3$ & $9.8 \pm 1.8$ & $9.4 \pm 2.1$ & $8.3 \pm 2.5$ \\ 
		\hline
	\end{tabular}
	\label{tbl:driftsAirports}
\end{table}

\begin{figure}[!ht]
	\centering
	\includegraphics[width=0.7\textwidth]{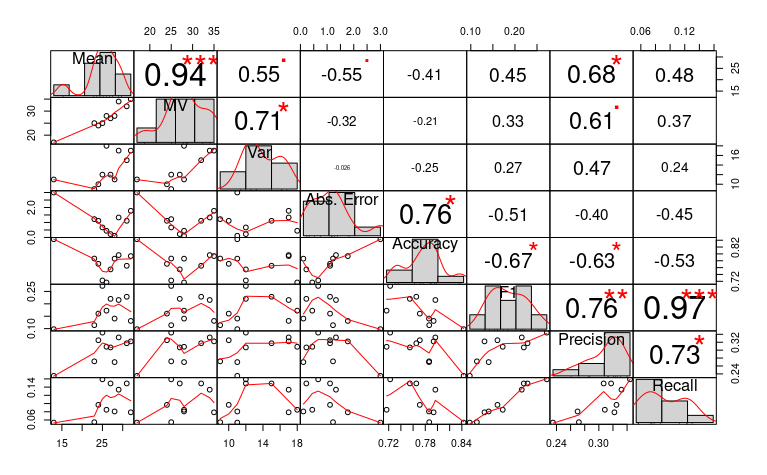}
	\caption{Correlation matrix among the number of drifts detected by $active$ drift handling strategies and prediction performance indicators.}
	\label{fig:correlation}
\end{figure}

To investigate the number of drifts, we tested the correlation among all performance indicators and drift detection methods. The $passive$ and $baseline$ had no drifts and were not considered on the number of drifts correlation. In this topic, the $recall$ was correlated with $accuracy$, shown in Figure \ref{fig:correlation}. Moreover, our experiments also showed that the number of drifts might be related to prediction performance. Specifically, the number of drifts of the drift detection method based on $mean$ comparison showed a significant positive correlation with $precision$. The drifts from $mean/variance$ method were also correlated to $precision$, which was expected considering the high correlation with $mean$ method. The drifts from the $mean$ method also showed high correlation values with $f_1$ and $recall$. These results indicate that the higher the frequency of drifts detected by $mean$, the higher are the $precision$ scores.

\begin{figure}[!ht]
	\centering
	\includegraphics[width=0.625\textwidth]{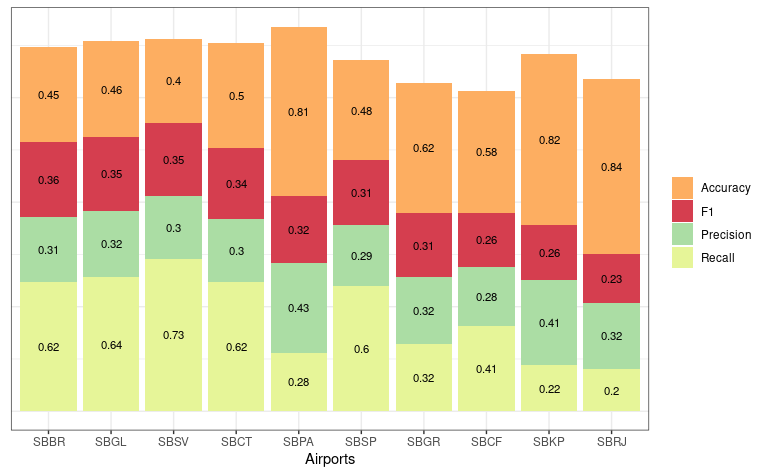}
	\caption{Prediction performance metrics for airports.}
	\label{fig:airportsperformance}
\end{figure}

For a more in-depth analysis, Figure \ref{fig:airportsperformance} presents the metrics for all airports. We also tested the correlation of the absolute error of each airport when compared to $SB$ number of detected drifts. The absolute error showed a high correlation with $accuracy$. In fact, $SBRJ$ and $SBCF$ showed the lowest number of drifts. They were also among the three lowest overall performance scores, as shown in Table \ref{tbl:driftsAirports} and Figure \ref{fig:correlation}. Besides, $SBKP$ is the one with the most average drifts. It showed the second lower score of $f_1$. These results indicate that extreme frequencies of drifts may impair prediction performance. Moreover, this can be related to the stability-plasticity dilemma \citep{haykin_neural_2011}, considering that too frequent or too rare retrains may impair prediction performance.

\subsection{Discussion}
\label{subsec:discussion}
Apart from the two main research questions, other important results were observed. 

\subsubsection*{Performance of Active Drift Detection Methods}

When analyzing the prediction performance of $active$ strategies, those based in $mean$ and $mean/variance$ drift detection methods may show better results than $variance$ alone for $AB$ scale. It may be explained by a high number of drifts in airports' data from the BFS dataset. As $variance$ method has shown, on average, half of $mean$ and $mean/variance$ tests, drifts that may decrease prediction performance would not imply model retraining, impairing results. However, the $variance$ method showed similar results to other $active$ drift detection methods for the top ten combinations of $SB$ scale.

\subsubsection*{Best $BSS$ and Classifiers}

\begin{figure}[!ht]
	\centering
	\includegraphics[width=0.7\textwidth]{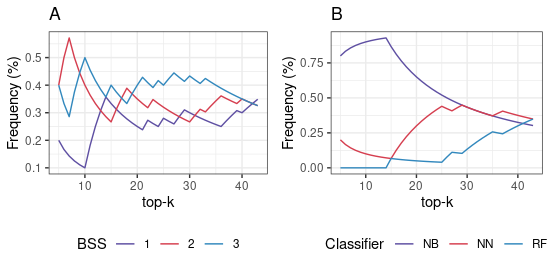}
	\caption{Frequency of each $BSS$ for each top-$k$ combination (A); Frequency of each Classifier for each top-$k$ combination for $SB$ analysis. (B)}
	\label{fig:top_k_bss_class}
\end{figure}

We also carried a top-$k$ analysis for $BSS$ and Classifiers. Regarding $BSS$ (Figure \ref{fig:top_k_bss_class}.a), in $SB$ scale, half of the top ten combinations used a $BSS$ of three years. Moreover, for $AB$ scale, 51.1\% of all top ten combinations from all airports had a three-year $BSS$. These results indicate a higher probability of achieving good prediction performances with a three-year $BSS$. It implies better prediction results when training and testing for drifts with more data from more extended periods.

Among classifiers (Figure \ref{fig:top_k_bss_class}.b), the scale seems to be related to prediction performance. Specifically, for the $SB$ scale, the $NB$ showed 90\% prevalence among the top ten combinations. However, for $AB$ scale, the most frequent classifier was $RF$ with 67\% prevalence, followed by $NN$ with 41\%, and $NB$ had only 2\% of all top ten combinations. First, considering that our hyperparameter optimization was only with the first batch, these results indicate that hyperparameters may have greater importance on $SB$ than on $AB$ scale. Moreover, for $AB$ scale, $RF$ may be more stable due to higher homogeneity of data, and $NN$ was more sensitive to hyperparameters tuning.

\subsubsection*{Comparison to other studies}

Regarding prediction performance, the present work showed similar $accuracy$ results to other works that investigated concept drifts in flight delay prediction \citep{kim_deep_2016, khamassi_drift_2014, pesaranghader_fast_2016}. However, as we expected, models that classify the entire data as a majority class may show a high $accuracy$, considering that the delay prevalence is around 20\%. Moreover, our experiments also showed that such an approach is not correlated with all prediction performance metrics. Specifically, $f_1$ and $precision$ were correlated with all other metrics, as shown in Figure \ref{fig:correlation}. The $accuracy$ was correlated with $f_1$ and $precision$, but not with $recall$. These results reinforce the importance of reporting multiple important prediction performance metrics in concept drift studies.

Another relevant discussion regards $BSS$. In our study, $BSS$ varied from one to three years were used. All other studies used hours and number of cases for defining $BSS$. In fact, the largest $BSS$ used for flight delay classification was of 9 hours \citep{kim_deep_2016}. Our approach made it possible to include seasonality when training classifiers.

\section{Conclusion} 
\label{sec:conclusion}

In this paper, we analyzed different types of drift handling strategies in aviation. Two research questions were answered to achieve the main objective of this study. It was observed that drift handling strategies are relevant. Their impact varies according to scale and machine learning models used. The experimental evaluation was done using a dataset that integrates weather and flight data from the Brazilian system.

In our analysis, the $passive$ and $active$ strategies together were more frequent among the top combinations for both scales tested ($AB$ and $SB$). It may be related to the high prevalence of drifts. In this case, strategies that retrain machine learning models offer better models than those that train only once.

The number of detected drifts on $AB$ analysis was related to prediction performance as well. Specifically, a high deviation from the number of drifts found in $SB$ was negatively correlated with better prediction performance metrics. It is related to the stability-plasticity dilemma, considering that too frequent or too rare retrains may impair prediction performance.

As limitations, the first one is those drift detection methods used in the experimental evaluation. They were focused only on changes $P(Y)$. The second limitation is related to hyperparameter optimizations, which were only established in the first $BSS$. Future studies may consider: (i) testing error-based drift handling strategies; (ii) investigating ensemble drift detection methods; (iii) performing novel hyperparameter optimizations once drift is observed; (iv) check drift handling strategies under different thresholds for flight delays; and (v) evaluate other metrics such as ROC curve.

\section*{Acknowledgments}
The authors thank CNPq, CAPES (finance code 001), FAPERJ, and CEFET/RJ for partially funding this research.

\section*{Conflict of interest}
On behalf of all authors, the corresponding author states that there is no conflict of interest.

\bibliographystyle{abbrvnat}
\bibliography{references}

\end{document}